\documentclass[a4paper]{article}
\usepackage[margin=25mm]{geometry}
\usepackage{url}
\usepackage{import}
\usepackage{amsfonts,amssymb,amsmath,amsthm,dsfont}
\usepackage{xcolor}
\usepackage{comment}
\usepackage[linesnumbered,algoruled]{algorithm2e}
\usepackage{bm}
\usepackage[normalem]{ulem}
\usepackage{graphicx}
\graphicspath{{figures/}}
\usepackage{adjustbox}
\usepackage{mathtools}

\providecommand{\keywords}[1]
{
  \small
  \textbf{\textit{Keywords---}} #1
}

\begin{document}

\title{A majorization-minimization algorithm for nonnegative binary matrix factorization\thanks{This work has received funding from the European Research Council (ERC) under the European Union’s Horizon 2020 research and innovation programme under grant agreement No 681839 (project FACTORY).}}
\date{}
\author{Paul~Magron\thanks{Université de Lorraine, CNRS, Inria, LORIA, F-54000 Nancy, France (e-mail: firstname.lastname@inria.fr).},
        Cédric~Févotte\thanks{IRIT, Université de Toulouse, CNRS, Toulouse, France (e-mail: firstname.lastname@irit.fr).}
}

\maketitle
\begin{abstract}
This paper tackles the problem of decomposing binary data using matrix factorization. We consider the family of mean-parametrized Bernoulli models, a class of generative models that are well suited for modeling binary data and enables interpretability of the factors. We factorize the Bernoulli parameter and consider an additional Beta prior on one of the factors to further improve the model's expressive power. While similar models have been proposed in the literature, they only exploit the Beta prior as a proxy to ensure a valid Bernoulli parameter in a Bayesian setting; in practice it reduces to a uniform or uninformative prior. Besides, estimation in these models has focused on costly Bayesian inference. In this paper, we propose a simple yet very efficient majorization-minimization algorithm for maximum a posteriori estimation. Our approach leverages the Beta prior whose parameters can be tuned to improve performance in matrix completion tasks. Experiments conducted on three public binary datasets show that our approach offers an excellent trade-off between prediction performance, computational complexity, and interpretability.
\end{abstract}

\section{Introduction}

Binary data are encountered in a variety of research fields such as paleontology~\cite{Bingham2009}, electoral data analysis~\cite{Voeten2013}, recommender systems~\cite{Schedl2015}, or binary image classification~\cite{Kaban2008}.

A popular approach for decomposing tabular data is matrix factorization (MF)~\cite{Koren2009}, which  consists in expressing a data matrix $\mathbf{Y}$ as the approximate product (of lower rank) of two matrices $\mathbf{W}$ and $\mathbf{H}$, along with constraints such as nonnegativity~\cite{Lee1999}. MF can be cast in a probabilistic framework~\cite{Salakhutdinov2007}, where the data is assumed to follow some distribution whose parameter is structured as $\mathbf{WH}$. Over the years, many distributions have been used in order to account for specific properties of the data at hand. For instance, Poisson generative models are appropriate for modeling count data encountered in recommender systems~\cite{Gopalan2014,Gopalan2015}. However, Poisson or Gaussian~\cite{Hu2008} models are commonly used to analyze binary data for practical reasons, even though they are not tailored for this task.

In order to explicitly account for the binary nature of the data, models based on the Bernoulli distribution have been proposed. These can be divided into two categories. On the one hand, models in the logistic principal component analysis (PCA)~\cite{Landgraf2020} family exploit a \textit{link} function in order to map the factorization to the space of Bernoulli parameters $[0, 1]$, i.e., $\mathbb{E}(\mathbf{Y}|\mathbf{WH}) = \sigma(\mathbf{WH})$. On the other hand, \textit{mean-parametrized} models~\cite{Lumbreras2020} directly factorize the Bernoulli parameter, i.e., $\mathbb{E}(\mathbf{Y}|\mathbf{WH}) = \mathbf{WH}$. Mean-parameterization is a useful property because it readily allows to interpret the factors and the approximation $\mathbf{WH}$. Nonetheless, to ensure a valid Bernoulli parameter, it is required to additionally constrain $\mathbf{W}$ and $\mathbf{H}$, e.g., using Dirichlet and Beta priors in a Bayesian setting~\cite{Lumbreras2020}. However, in these approaches the Beta prior's sole purpose it to serve as a proxy to ensure a valid parameter: in practice it simplifies to a uniform or uninformative prior~\cite{Kaban2008, Lumbreras2020}, thus its full potential remains to be assessed. Besides, variational~\cite{Kaban2008} or sampling~\cite{Lumbreras2020} estimation schemes are computationally costly, as pointed out in~\cite{Lumbreras2020}. Finally, the expectation-maximization (EM) algorithm from~\cite{Bingham2009} does not use any prior, and relies on an augmented model with hidden variables, which somehow complicates the derivations.

In this paper, we propose a new approach for estimating a mean-parametrized Bernoulli model, which alleviates the aforementioned issues. We summarize hereafter our contributions and their advantages:
\begin{enumerate}
    \item We consider a Beta prior on $\mathbf{H}$ with tunable hyperparameters: this allows to optimally exploit this prior, whose impact on performance was until now left to explore.
    \item We consider maximum a posteriori (MAP) estimation of the model's parameters with majorization-minimization (MM)~\cite{Hunter2004,Sun2017}. This yields easy-to-implement updates with no extra parameters (such as step sizes) to adjust. It is considerably more straightforward to derive than previous approaches. It notably generalizes the prior-free EM method from~\cite{Bingham2009}, without using  hidden variables and data augmentation.
    \item We compare our method with state-of-the-art logistic PCA for matrix completion. We show that our method exhibits an excellent trade-off between prediction performance, computational complexity, and interpretability.
\end{enumerate}
The rest of this paper is structured as follows. Section~\ref{sec:models} introduces the generative model and underlines its connection with related works. Section~\ref{sec:method} presents the MAP estimation procedure with MM. Experiments are reported in Section~\ref{sec:exp}. Finally, Section~\ref{sec:conclu} draws some concluding remarks.

\vspace{0.5em}

\noindent \textbf{Mathematical notations}:

\begin{itemize}
    \item $a$ (regular): scalar.
    \item $\mathbf{a}$ (lower case, bold font): vector.
    \item $\mathbf{A}$ (capital, bold font): matrix. The $(m,n)$-th entry of $\mathbf{A}$ is denoted $[\mathbf{A}]_{m,n} = a_{m,n}$.
\end{itemize}

\section{Binary data models}
\label{sec:models}

In this section we briefly present Bernoulli-based MF models for binary data. For a more detailed overview, we refer the interested reader to Table 1 from~\cite{Lumbreras2020}.

\subsection{Logistic PCA}

Let us consider a binary data matrix $\mathbf{Y} \in \{ 0, 1 \}^{M \times N}$. Logistic PCA builds upon the following generative model:
\begin{equation}
    y_{m,n} \sim \text{Bernoulli}(\sigma([\mathbf{W} \mathbf{H}]_{m,n})),
    \label{eq:bernoulli_link}
\end{equation}
where $\mathbf{W} \in \mathbb{R}^{M \times K}$, $\mathbf{H} \in \mathbb{R}^{K \times N}$, $K$ is the rank of the factorization, and $\sigma: x \to 1/(1+e^{-x})$ is the logistic function,\footnote{Note that alternative link functions have been considered, e.g., in~\cite{Zhou2015}.} which maps the factorization to the range $[0,1]$. Logistic PCA has been computed using a variety of techniques, including variational approaches~\cite{Tipping1998}, gradient descent~\cite{Collins2001}, and alternating least squares~\cite{Schein2003}.

Thanks to the mapping function, no constraint is needed on the factors to ensure a valid Bernoulli parameter. Nonetheless, several approaches have additionally enforced the nonnegativity of one~\cite{Tome2015} or two~\cite{Larsen2015} factors, or leveraged Gaussian priors~\cite{Tipping1999,Ortega2021}. Despite its popularity and performance, a drawback of logistic PCA stems from the fact that the link function hampers interpretability of the decomposition, which is often a desired feature, e.g., for analyzing econometric data~\cite{Hidalgo2021}.

\subsection{Mean-parametrized Bernoulli models}
\label{sec:models_meanparam}

To alleviate the aforementioned issue, mean-parametrized Bernoulli models~\cite{Lumbreras2020} have been proposed, such that:
\begin{equation}
    y_{m,n} \sim \text{Bernoulli}([\mathbf{W} \mathbf{H}]_{m,n}).
    \label{eq:bernoulli}
\end{equation}
In order to guarantee that $[\mathbf{W} \mathbf{H}]_{m,n} \in [0,1]$, it is mandatory to impose additional constraints on the factors, such as:
\begin{equation}
    \forall m, \sum_k w_{m,k} = 1 \quad \text{and} \quad \forall (k,n), h_{k,n} \leq 1,
    \label{eq:constraints_bern}
\end{equation}
along with the nonnegativity of $\mathbf{W}$ and $\mathbf{H}$.\footnote{Other constraints are possible, as will be detailed in Section~\ref{sec:method_algo}.} As a result, we call this family of models NBMF, which stands for \textit{nonnegative binary matrix factorization}. Maximum likelihood estimation in a such a model was proposed in~\cite{Bingham2009} by introducing hidden variables and deriving an EM algorithm, yielding a variant that we name NBMF-EM. In a Bayesian setting, it is common to consider the following priors instead of~\eqref{eq:constraints_bern}:
\begin{equation}
    \mathbf{w}_m \sim \text{Dirichlet}(\boldsymbol{\gamma}) \quad \text{and} \quad h_{k,n} \sim \text{Beta}(\alpha_{k}, \beta_{k}),
    \label{eq:constraints_priors}
\end{equation}
where $\mathbf{w}_m$ denotes the $m$-th row of $\mathbf{W}$. Such models have been estimated using variational Bayesian approaches~\cite{Kaban2008} or collapsed Gibbs sampling~\cite{Lumbreras2020}. In these approaches, the priors only serve as proxy to ensure valid Bernoulli parameters: in practice the Beta prior reduces to a uniform (${\alpha_k=\beta_k=1}$)~\cite{Lumbreras2020} or uninformative (${\alpha_k=\beta_k=1/2}$)~\cite{Kaban2008} prior. Therefore, these approaches have not actually assessed the potential of carefully tuning the parameters of this prior. We will show that tuning these parameters can lead to significant improvements.

\section{Proposed method}
\label{sec:method}

Let us consider the NBMF generative model with the Beta prior in \eqref{eq:constraints_priors} for $\mathbf{H}$ and the sum-to-one constraint in \eqref{eq:constraints_bern} for $\mathbf{W}$. We now derive our MM algorithm for MAP estimation. 

\subsection{Objective}
\label{sec:method_problem}

We seek to minimize $f(\mathbf{W}, \mathbf{H}) + g(\mathbf{H})$ under the constraints~\eqref{eq:constraints_bern}, where $f$ is the negative Bernoulli log-likelihood:
\begin{equation}
     f(\mathbf{W}, \mathbf{H}) = - \log p(\mathbf{Y}|\mathbf{WH})  =  - \sum_{m,n}  y_{m,n} \log([\mathbf{W} \mathbf{H}]_{m,n})   + (1 - y_{m,n}) \log(1 - [\mathbf{W} \mathbf{H}]_{m,n}), \label{eq:likelihood}
\end{equation}
and $g$ is the negative Beta log-prior:
\begin{equation}
     g(\mathbf{H}) = - \log p(\mathbf{H}) = - \sum_{k,n} (\alpha_{k}-1) \log(h_{k,n}) + (\beta_{k}-1) \log(1-h_{k,n}).
    \label{eq:log_beta}
\end{equation}
We account for the constraint on $\mathbf{W}$ via the method of Lagrange multipliers (denoted $\lambda_m$), thus the problem becomes that of finding a stationary point for:
\begin{equation}
    \mathcal{L}(\mathbf{W},\mathbf{H}, \boldsymbol{\lambda}) = f(\mathbf{W}, \mathbf{H}) + g(\mathbf{H}) + \sum_m \lambda_m \left( \sum_k w_{m,k} - 1 \right).
    \label{eq:L}
\end{equation}
Note that we do not explicitly consider the constraint $h_{k,n} \leq 1$ nor the nonnegativity of the factors, since we will prove in Section~\ref{sec:method_algo} that these automatically hold in our algorithm.

\subsection{Estimation with majorization-minimization}
\label{sec:method_mm}

We consider MM~\cite{Sun2017}, which has shown powerful for estimating MF models in many settings~\cite{Fevotte2011,Magron2018levywaspaa,Lin2018}. In a nutshell, if we consider minimization of a function $\phi$ with parameters $\theta$ and current estimate $\tilde{\theta}$, MM consists in constructing and minimizing a tight upper bound $\psi$ such that:
\begin{equation}
   \forall \theta, \quad \phi(\theta) \leq \psi(\theta, \tilde{\theta}) \quad \text{and} \quad \phi(\tilde{\theta}) = \psi(\tilde{\theta}, \tilde{\theta}).
\end{equation}
Then, it can easily be shown~\cite{Hunter2004} that $\phi$ is non-increasing under the following update scheme: $\tilde{\theta} \leftarrow \arg \min_{\theta} \psi(\theta, \tilde{\theta})$. Here we consider a block-descent strategy in which $\mathbf{H}$ and $\mathbf{W}$ are updated in turn, which produces a valid descent algorithm.

\subsubsection{Update on $\mathbf{H}$}

Let us first present the update for $\mathbf{H}$. To that end, we seek to majorize $f$ defined in~\eqref{eq:likelihood} with respect to $\mathbf{H}$, with $\mathbf{W}$ fixed. Denoting the current estimate by $\tilde{\mathbf{H}}$, the first term in~\eqref{eq:likelihood}, denoted $f_1(\mathbf{H})$, can be rewritten as:
\begin{equation}
    \log \left( \sum_{k} w_{m,k} h_{k,n} \right) = \log \left( \sum_{k} \tilde{\rho}_{m,n,k} \frac{w_{m,k}}{h_{k,n} \tilde{\rho}_{m,n,k} } \right),
     \label{eq:majH_firstterm}
\end{equation}
where ${\tilde{\rho}_{m,n,k} = w_{m,k} \tilde{h}_{k,n} / \tilde{y}_{m,n}}$, and $\tilde{y}_{m,n} = \sum_l w_{m,l} \tilde{h}_{l,n}$. Since the weights $\tilde{\rho}_{m,n,k}$ are nonnegative and $\sum_k \tilde{\rho}_{m,n,k} =1$, and since the function ${x \to -\log x}$ is convex, we can majorize $f_1$ using Jensen inequality, such that $f_1(\mathbf{H}) \leq \psi_1(\mathbf{H}, \tilde{\mathbf{H}}) $ with:
\begin{equation}
     \psi_1(\mathbf{H}, \tilde{\mathbf{H}}) = - \sum_{m,n,k} \frac{y_{m,n} w_{m,k} \tilde{h}_{k,n}}{\tilde{y}_{m,n}} \log \left( \frac{h_{k,n} \tilde{y}_{m,n}}{\tilde{h}_{k,n} } \right).
     \label{eq:psi1}
\end{equation}
To obtain a majorization for the second term $f_2(\mathbf{H})$ in~\eqref{eq:likelihood}, a first naive approach consists in exploiting the convexity of ${x \to -\log (1-x)}$ similarly as above. However, this leads to an intractable minimization step. Instead, we exploit the constraint $\sum_k w_{m,k} = 1$ to rewrite $f_2(\mathbf{H})$ as follows:
\begin{align}
     \log(1 - \sum_k & w_{m,k}  h_{k,n}) =  \log(\sum_k w_{m,k} (1 - h_{k,n})) \\
     &= \log \left( \sum_k \tilde{\mu}_{m,n,k} \frac{w_{m,k} ( 1 - h_{k,n})}{\tilde{\mu}_{m,n,k} } \right)
\end{align}
where we have introduced the nonnegative weights $\tilde{\mu}_{m,n,k} = w_{m,k} (1-\tilde{h}_{k,n}) / (1-\tilde{y}_{m,n})$, which also sum up to~$1$. Using again Jensen inequality, we obtain an upper bound for this second term, i.e., $ f_2(\mathbf{H}) \leq \psi_2(\mathbf{H}, \tilde{\mathbf{H}}) $ with:
\begin{equation}
     \psi_2(\mathbf{H}, \tilde{\mathbf{H}}) = - \sum_{m,n,k} \frac{(1-y_{m,n}) w_{m,k} (1-\tilde{h}_{k,n})}{1-\tilde{y}_{m,n}} \times \log \left( \frac{ (1-h_{k,n}) (1-\tilde{y}_{m,n})}{1-\tilde{h}_{k,n}} \right).
     \label{eq:psi2}
\end{equation}
Combining~\eqref{eq:psi1} and~\eqref{eq:psi2} leads to $f \leq \psi_1 + \psi_2$, which yields an upper bound of the Lagrangian $\mathcal{L}$ given by \eqref{eq:L}. Note that the upper bound is tight, since it is straightforward to prove that equality holds when $\mathbf{H}=\tilde{\mathbf{H}}$. Minimizing the upper-bound (which is separable, smooth and convex) results in
\begin{equation}
    h_{k,n} = \displaystyle \frac{\tilde{c}_{k,n}}{\tilde{c}_{k,n}+\tilde{d}_{k,n}},
    \label{eq:update_H}
\end{equation}
where
\begin{align}
    \tilde{c}_{k,n} &= \tilde{h}_{k,n} \sum_m  \frac{y_{m,n} w_{m,k}}{\tilde{y}_{m,n}} +  \alpha_{k} - 1, \\
    \tilde{d}_{k,n} &= (1-\tilde{h}_{k,n}) \sum_m  \frac{(1-y_{m,n}) w_{m,k}}{1-\tilde{y}_{m,n}} + \beta_{k}-1.
    \label{eq:C_D}
\end{align}

%
%
%
%
%
%

\subsubsection{Update on $\mathbf{W}$} An upper bound of $\mathcal{L}$ w.r.t to $\mathbf{W}$ for fixed $\mathbf{H}$ and $\boldsymbol{\lambda}$ can be obtained using the same tricks as above. Canceling the gradient of the upper bound now leads to:
\begin{equation}
    - \frac{\tilde{w}_{m,k}}{w_{m,k}} \left( \sum_n  \frac{y_{m,n} h_{k,n}}{\tilde{y}_{m,n}} + \frac{(1-y_{m,n}) (1-h_{k,n})}{1-\tilde{y}_{m,n}}  \right)  + \lambda_m = 0.
    \label{eq:gradM_W}
\end{equation}
To determine $\lambda_m$, we multiply~\eqref{eq:gradM_W} by $w_{m,k}$ and sum over $k$ to exploit the constraint~\eqref{eq:constraints_bern}. This yields:
\begin{equation}
    \lambda_m  = \sum_n  \frac{y_{m,n}}{\tilde{y}_{m,n}} \sum_k \tilde{w}_{m,k} h_{k,n}  + \frac{1-y_{m,n}}{1-\tilde{y}_{m,n}} \sum_k \tilde{w}_{m,k} (1 - h_{k,n}).
    \label{eq:gradM_W_lambda}
\end{equation}
Since ${\sum_k \tilde{w}_{m,k} (1 - h_{k,n}) = 1 - \tilde{y}_{m,n}}$ and $\sum_k \tilde{w}_{k,m} h_{k,n} = \tilde{y}_{m,n}$, the expression of $\lambda_m$ simplifies to:
\begin{equation}
    \lambda_m  = \sum_n y_{m,n} + (1-y_{m,n}) = N.
    \label{eq:gradM_W_lambda_simp}
\end{equation}
Finally, combining~\eqref{eq:gradM_W} and~\eqref{eq:gradM_W_lambda_simp} yields the following update: 
\begin{equation}
    w_{m,k} = \tilde{w}_{m,k} \left( \sum_n \frac{y_{m,n} h_{k,n}}{\tilde{y}_{m,n}} + \frac{(1-y_{m,n}) (1-h_{k,n})}{1-\tilde{y}_{m,n}} \right) / N.
    \label{eq:update_W}
\end{equation}

\subsection{Algorithm}
\label{sec:method_algo}

Alternating~\eqref{eq:update_H} and~\eqref{eq:update_W} leads to the iterative procedure that we name NBMF-MM. It is summarized in Algorithm~\ref{al:NBMF}, where the updates are written into matrix form. The operations $.^\mathsf{T}$, $\odot$, and $\frac{\cdot}{\cdot}$ denote matrix transpose, element-wise multiplication, and division, respectively.
Note that Algorithm~\ref{al:NBMF} uses constant hyperparameter values $\alpha_{k} = \alpha$ and $\beta_{k} = \beta$, which led to satisfactory performance in preliminary experiments.

\begin{algorithm}[t]
	\caption{NBMF-MM}
	\label{al:NBMF}
			\textbf{Inputs}: Data matrix $\mathbf{Y} \in \{0, 1\}^{M \times N}$, prior parameters $\alpha \geq 1$ and $\beta \geq 1$  \\

            \textbf{Initialize} $\mathbf{W}$ and $\mathbf{H}$ such that they comply with~\eqref{eq:constraints_bern}. \\
            
            \While{convergence not reached}{
            $ \mathbf{C} = \mathbf{H} \odot \left( \mathbf{W}^{\mathsf{T}} \displaystyle \frac{\mathbf{Y}}{\mathbf{W}\mathbf{H}} \right) + \alpha - 1  $ \\
            $ \mathbf{D} = (1-\mathbf{H}) \odot \left( \mathbf{W}^{\mathsf{T}} \displaystyle \frac{1-\mathbf{Y}}{1-\mathbf{W}\mathbf{H}}\right) + \beta - 1 $ \\
            $ \mathbf{H} = \displaystyle \frac{\mathbf{C}}{\mathbf{C} + \mathbf{D}}$ \\
            $ \mathbf{W} =  \mathbf{W} \odot \left( \displaystyle \frac{\mathbf{Y}}{\mathbf{W}\mathbf{H}} \mathbf{H}^{\mathsf{T}}  + \displaystyle \frac{1 - \mathbf{Y}}{1 - \mathbf{W}\mathbf{H}} (1 - \mathbf{H})^{\mathsf{T}} \right) / N  $
            }
			\textbf{Outputs}: $\mathbf{W}$, $\mathbf{H}$
\end{algorithm}

We remark that if $\mathbf{W}$ and $\mathbf{H}$ are initialized with nonnegative entries that respect the constraints~\eqref{eq:constraints_bern}, then the proposed updates guarantee that these constraints hold through iterations. Indeed, since initially $0 \leq \mathbf{H} \leq 1$ and $0 \leq \mathbf{WH} \leq 1$, then $1-\mathbf{H} \geq 0$ and $1 - \mathbf{WH} \geq 0$. Besides, since $\mathbf{Y}$ is binary, then $1 - \mathbf{Y} \geq 0$. Therefore, all the terms involved in the updates are nonnegative, and consequently $\mathbf{W}$ and $\mathbf{H}$ remain nonnegative. Moreover, since the update on $\mathbf{H}$ is of the form $\mathbf{C} / (\mathbf{C} + \mathbf{D})$ with $\mathbf{C}$ and $\mathbf{D}$ nonnegative (as long as $\alpha \geq 1$ and $\beta \geq 1$), then $\mathbf{H} \leq 1$. Thus, the constraints are preserved.

Let us point out that other sets of constraints can ensure a valid Bernoulli parameter~\cite{Lumbreras2020}. Indeed, one can switch the role of $\mathbf{W}$ and $\mathbf{H}$ (and the constraints/priors accordingly), which results in switching the corresponding updates in Algorithm~\ref{al:NBMF}. Alternatively, it is possible to set $\sum w_{m,k} =1$ and $\sum_k h_{k,n} = 1$, in which case the update on $\mathbf{H}$ becomes similar to that of $\mathbf{W}$. In this work we only consider~\eqref{eq:constraints_bern} for brevity.

Finally, let us outline that if the Beta prior reduces to a uniform prior (i.e., setting $\alpha=\beta=1$), then the procedure is equivalent to the NBMF-EM algorithm~\cite{Bingham2009}. However, NBMF-MM is obtained in a more straightforward fashion thanks to the MM strategy, does not require to introduce latent variables in an augmented model, and allows to tune the prior parameters.

\section{Experiments}
\label{sec:exp}

In this section, we assess the potential of NBMF-MM for decomposing and predicting binary data in a matrix completion task. Our code is available online for reproducibility.\footnote{The code will be made available when the paper is published.}

\subsection{Protocol}

\subsubsection{Datasets}

We consider three public binary datasets:
\begin{itemize}
    \item \texttt{animals}~\cite{Kemp2006}: An entry ${y_{m,n}=1}$ indicates that the animal $m$ has the attribute $n$ ($M=50$, $N=85$).
    \item \texttt{paleo}~\cite{Bingham2009}: An entry $y_{m,n}=1$ indicates that the gene $m$ has been found at location $n$ ($M=253$, $N=902$).
    \item \texttt{lastfm}~\cite{Bertin2011}: An entry $y_{m,n}=1$ indicates that the user $m$ has listened to the artist $n$ ($M=1,226$, $N=285$).
\end{itemize}
Each dataset is split into a training, a validation, and a test subset, containing $70 \%$, $15 \%$ and $15 \%$ of the data, respectively. The factors are learned on the training subset, and the hyperparameters (rank of the factorization $K$ and prior parameters $\alpha$ and $\beta$) are tuned to minimize perplexity (see below) on the validation subset. Finally, the trained model is used to predict the test data in a binary matrix completion task.

\subsubsection{Methods} The proposed NBMF-MM is compared against two baselines: NBMF-EM~\cite{Bingham2009}, which is equivalent to Algorithm~\ref{al:NBMF} with $\alpha=\beta=1$; and logistic PCA (logPCA)~\cite{Landgraf2020}, which is a state-of-the-art method for predicting binary data.\footnote{We use the package available at \url{https://github.com/andland/logisticPCA}} The Bayesian methods from~\cite{Kaban2008} and~\cite{Lumbreras2020} are also relevant baselines, but they have been shown to perform similarly to NBMF-EM on these datasets~\cite{Lumbreras2020}. Thus, for brevity we do not report these. All methods use the same convergence criterion: the algorithm is stopped when the relative variation of the objective function is lower than $10^{-5}$ or when a maximum number of $2000$ iterations is reached.

\subsubsection{Evaluation}
Predictions are computed through ${\hat{\mathbf{Y}} = \mathbf{W} \mathbf{H}}$ for NBMF models, and $\hat{\mathbf{Y}} = \sigma(\mathbf{W} \mathbf{H})$ for logPCA. Prediction performance is then measured using the \textit{perplexity}~\cite{Hofmann1999} (lower is better), which is defined as:
\begin{equation}
    \text{perplexity} = - \frac{1}{|\vartheta|} \sum_{(m,n) \in \vartheta} \log p(y_{m,n} | \hat{y}_{m,n} )
\end{equation}
where $\vartheta$ denotes the evaluation (validation or test) set, and $|\vartheta|$ denotes the number of elements in $\vartheta$. 

\subsection{Results}

\begin{figure}[t]
    \centering
    \includegraphics[width=.6\linewidth]{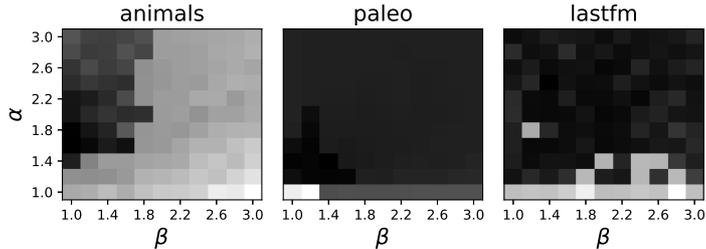}
    \vspace{-2em}
    \caption{Validation perplexity (darker is better) for the optimal rank $K$.}
    \label{fig:val_alpha_beta}
\end{figure}

\begin{figure}[t]
    \centering
    \includegraphics[width=.6\linewidth]{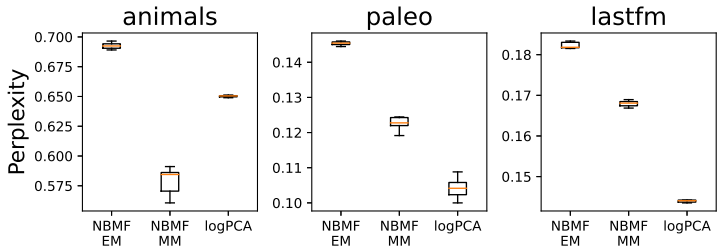}
    \vspace{-2em}
    \caption{Perplexity of the test set for $10$ random initializations. Each box-plot is made up of a central line indicating the median, box edges indicating the $1^{\text{st}}$ and $3^{\text{rd}}$ quartiles, and whiskers indicating the extremal values.}
    \label{fig:test_pplx}
\end{figure}

\begin{figure}[t]
    \centering
    \includegraphics[width=.7\linewidth]{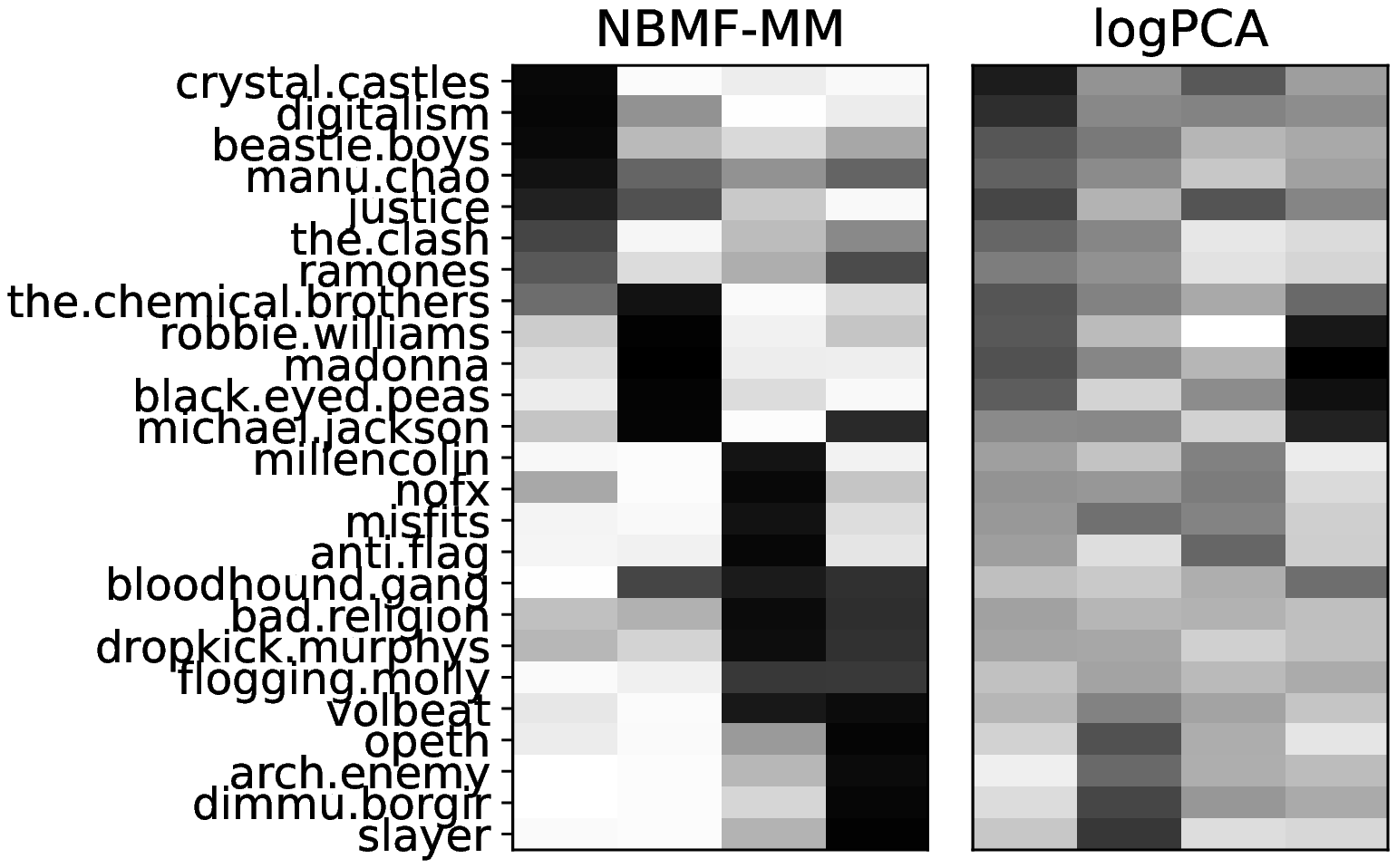}
    \vspace{-2em}
    \caption{Estimated $\mathbf{H}$ matrix (transposed) from the \texttt{lastfm} dataset.}
    \label{fig:H_lastfm}
\end{figure}

First, we investigate the impact of the Beta prior on the performance on NBMF-MM. We display the perplexity on the validation set in Fig.~\ref{fig:val_alpha_beta}. Overall, while increasing $\alpha$ improves performance (up to a point which depends on the dataset), a different trend is observed regarding $\beta$. For instance, performance becomes worse than with prior-free NBMF for large values of $\beta$ and $\alpha=1$ on the \texttt{animals} dataset. On the other hand, performance on the \texttt{lastfm} dataset is less sensitive to these variations, provided that $\alpha > 1$.
We select the optimal hyperparameters and display the results on the test set in Fig.~\ref{fig:test_pplx}.
NBMF-MM outperforms NBMF-EM by a large magin, which demonstrates the effectiveness of adjusting the Beta prior. This also shows that similar methods, e.g.,~\cite{Lumbreras2020}, which have considered such a prior in the model formulation but did not test it experimentally, could actually benefit from this finding in order to fully reveal their potential. NBMF-MM outperforms logPCA on the \texttt{animals} dataset, but logPCA yields the best performance on \texttt{paleo} and \texttt{lastfm}. However, this result can be tempered by the following considerations. Firstly, NBMF-MM is roughly $10$ times faster than logPCA. Secondly, logPCA relies on using a link function, which hampers its interpretability.

To illustrate this last point, we plot in Fig.~\ref{fig:H_lastfm} the matrices $\mathbf{H}$ obtained on \texttt{lastfm}. We observe that much more distinct clusters can be extracted from the NBMF-based factor, which allows to grasp a high-level meaning of the $K$ components. On this example, we can indeed interpret these components as related to the musical genre, e.g., ``pop", ``electronic", ``rock" and ``punk/metal". This property is an asset in scenarios where interpretability of the factors is required.

\section{Conclusion}
\label{sec:conclu}

We have proposed a novel MF algorithm for decomposing binary data. This method builds upon a mean-parametrized Bernoulli generative model along with a Beta prior. The parameters are estimated with MM, which yields simple and efficient updates. Our method offers an excellent trade-off between prediction performance, computational complexity, and interpretability, compared to state-of-the-art logistic PCA. 

\newpage
\bibliographystyle{IEEEtran}
\bibliography{references}

\end{document}